# Fuzzy Longest Common Subsequence Matching With FCM

Ibrahim Ozkan[1][i,a], I. Burhan Türkşen [ii,b,c]

[a]Hacettepe Univ. Ankara, Turkey
[b]TOBB ETU, Ankara, Turkey
[c]Department of Mechanical and Industrial Engineering,
University of Toronto, Ontario, M5S3G8, Canada
[i]ozkan@mie.utoronto.ca, [ii]turksen@mie.utoronto.ca, bturksen@etu.edu.tr

ABSTRACT

Capturing the interdependencies between real valued time series can be achieved by finding common similar patterns. The abstraction of time series makes the process of finding similarities closer to the way as humans do. Therefore, the abstraction by means of a symbolic levels and finding the common patterns attracts researchers. One particular algorithm, Longest Common Subsequence, has been used successfully as a similarity measure between two sequences including real valued time series. In this paper, we propose Fuzzy Longest Common Subsequence matching for time series.

## INTRODUCTION

The knowledge discovery from the time series is at the core of active research in many fields. Pattern recognition is a particular field that is dealing with finding useful information from time series (in general, from sequences). In many settings, gathering useful information may include finding interdependencies between time series. Although, the concept of interdependency is more theoretical and context dependent (Batista et al, 2011), several similarity measures that are suitable for assessing the interdependency have been created and discussed in the literature. Among them, correlation, Euclidian distance, brownian distance correlation (Szekely et al. 2007, Szekely & Rizzo, 2009, Szekely & Rizzo, 2010), maximal information (Reshef et al. 2011), Markov

---

[1] Corresponding author.
E-mail addresses: ozkan@mie.utoronto.ca, turksen@mie.utoronto.ca

Operator Distance (Gregorio & Iacus, 2010), mutual information, Dynamic Time Warping (DTW), and the longest common subsequence (LCS) distances are worthy of mention. One can find simple explanatory examples of some quantitative measures that fail to capture the similarity between time series in literature (see for example, Hoppner, 2002a, Hoppner, 2002b, Jachner et al. 2007). It appears that the extracting knowledge from time series needs to mimic the way as human mind does it.

Knowledge is not revealed by numbers but rather it is built in human mind with the help of numbers, figures and objects (Hoppner, 2002a, Hoppner, 2002b). Summary measures, graphical representations are very useful tools to reveal some insight from data in general. There are numerous time series representation schemes suggested in the literature that help us to understand some properties (characteristics, features) of data. Some of them are high level representations that are created to overcome speed and quality issues for the properties of time series. Fourier Transform (representing time series with best 5-10 frequencies), wavelets (to capture time-frequency space properties), eigenwaves, local polynomial models, piecewise linear approximations, local trend information, SAX, etc., can be listed as related algorithms. (Goodrich, 1994, Imai & Iri 1986, Keogh & Smyth, 1997, Lin et al. 2007, Wang et al. 2006).

In some cases, many numerical simple summary and distance measures may be misleading. Quick measures such as Euclidian distance, correlation, structural characteristics, etc., measures have poor performances for time series which are: i) very noisy, ii) containing several outliers, iii) have position of the patterns which are not synchronized, iv) containing stretching/relaxing patterns (Vlachos et al. 2002, Vlachos et al. 2003). Therefore, capturing the similarities in a more abstract way as humans do is a central work in many knowledge discovery algorithms. Hoppner (2002b) suggests three steps to analyze interdependencies. First step is labeling (simple abstraction) real valued time series (or describing the patterns that the series contain as, "convex", "concave", "convex-concave", "concave-convex", etc.). This task is usually performed by means of some algorithms suitable for time series abstractions that may include some forms of expert knowledge, rules of thumb or clustering algorithms. The second step is finding the common patterns with suitable algorithms. The last step is deriving rules about pattern dependencies.

This article is focusing on the first two steps in fuzzy settings which results in a novel dissimilarity measure. The abstraction step is performed with fuzzy clustering (fuzzy c-means, FCM, in particular) and the common patterns are found with an application of LCS algorithm. This novel algorithm may be called as Fuzzy LCS which can be seen as the extension of LCS with an application of fuzzy calculations for real valued time series. Fuzzy LCS is briefly explained in section 2 and section 3. Then this algorithm is applied to artificially generated random walk series, sine function as for deterministic real valued series and some real world time series, such as, foreign exchange series and oil prices. Finally in section 4, conclusions are stated.

## Longest Common Subsequence with Fuzzy Sets, Fuzzy LCS

Labeling the time series such as "increasing", "decreasing", "convex", "concave", naturally includes some uncertainties since these words do not precisely describe any quantity. This type of abstraction can be achieved successfully with an application of fuzzy logic. Hence, the first step of the algorithm is to obtain fuzzy sets for real valued time series with an application of FCM algorithm. Once the Fuzzy Sets are obtained, can compare/match them by means of fuzzy operators. The second step of the algorithm is to obtain the longest common fuzzy subsequences with an application of LCS algorithm that uses fuzzy matching of fuzzy sequences. Thus, FCM algorithm is briefly explained first. Then the LCS algorithm is introduced and finally the fuzzy LCS algorithm is presented in this section.

## Fuzzy C-Means

Fuzzy C-Means (FCM) algorithm (Bezdek, 1973) partitions data into clusters in which each observation is assigned a membership value between zero and one to each cluster based on the minimization of the following objective function:

$$J_m(U,V:X) = \sum_{k=1}^{nd} \sum_{c=1}^{nc} \mu_{c,k}^m \|x_k - v_c\|_A^2 \qquad (1)$$

where, $\mu_{c,k}$ is membership value of kth vector in cth cluster such that $\mu_{c,k} \in [0,1]$, nd is the number of vectors used in the analysis, nc is the number of clusters, $\|.\|_A$ is norm, e.g., Euclidian or Mahalanobis, and m is the level of fuzziness, the membership values are calculated as:

$$\mu_{c,k} = \left[ \sum_{i=1}^{nc} \left( \frac{\|x_k - v_c\|_A}{\|x_k - v_i\|_A} \right)^{\frac{2}{m-1}} \right]^{-1} \qquad (2)$$

where, $\sum_{c=1}^{nc} \mu_{c,k} = 1$ for some given m>1, and finally the cluster centers are computed as:

$$v_c = \frac{\sum_{k=1}^{nd} \mu_{c,k}^m x_k}{\sum_{k=1}^{nd} \mu_{c,k}^m} \quad (3)$$

In fuzzy clustering analysis, the number of clusters, nc, and the level of fuzziness, m, need to be identified before clustering. There are several validation indices proposed for the number of clustersi. The number of cluster can be set based on expert knowledge or by means of cluster validity measure(s). There are limited studies for the assignment of the value of the level of fuzziness (Ozkan &Turksen, 2004, Ozkan &Turksen, 2007, Pal & Bezdek, 1995, Yu et al. 2004). The most widely used value is two (rule of thumb). In some cases researchers seek the optimal value based on some performance measure. Lately, Ozkan and Turksen (2007) suggest a value between upper and lower level of fuzziness which are calculated as 1.4 and 2.6 respectively.

## Longest Common Subsequence

Longest Common Subsequence is a subsequence, S, of the maximal length between two strings, say A and B. Let, $S = s_1, s_2, ..., s_p$ is a subsequence of both $A = a_1, a_2, ...., a_n$ and $B = b_1, b_2, ...., b_m$ where $p \prec m \leq n$. Then the mappings are defined as, $F_A : \{1,2,...,p\} \to \{1,2,...,n\}$ and $F_B : \{1,2,...,p\} \to \{1,2,...,m\}$ such that $F_A(i) = j$ if $s_i = a_j$ (similarly $F_B(i) = j$ if $s_i = b_j$) and mapping functions are monotone strictly increasing (Bergroth et al. 2000, Hirschberg, 1977, Lawrence & Wagner, 1975, Sellers, 1974, Wagner, 1975). It is then easy to compute the similarity between two strings directly related with the length of LCS. The degree of similarity is increasing with the length of LCS.

The following measure is copied and adapted from Vlachos et al. (2002) which is used to adapt LCS for real valued time series. Given an integer $\delta$ and a real number $0 < \epsilon < 1$, the distance , $D_{\delta,\epsilon}$, between to time series $A$ and $B$ with lengths of $m$ and $n$ respectively is defined as:

$$D_{\delta,\epsilon}(A,B) = 1 - \frac{LCS_{\delta,\epsilon}(A,B)}{\min(m,n)}$$

where

$$LCS_{\delta,\epsilon}(A,B) = \begin{cases} 0 & \text{if } A \text{ or } B \text{ is empty} \\ 1 + LCS_{\delta,\epsilon}(Head(A), Head(B)) & \text{if } |a_n - b_m| < \epsilon \text{ and } |n-m| \leq \delta \\ \max(LCS_{\delta,\epsilon}(Head(A),B), LCS_{\delta,\epsilon}(A,Head(B))) & \text{otherwise} \end{cases} \quad (4)$$

and $Head(A) = (a_1, a_2,..., a_{n-1})$, $Head(B) = (b_1, b_2,..., b_{m-1})$.

There are two parameters to be set before measuring the LCS similarity as a distance. These are integer $\delta$ which controls lag/lead time and a real number $\in$ where the sequences are treated as very close if the absolute value of difference between them is less then this value. It is also possible to use different lag and lead time parameters ($\delta_u$, $\delta_l$). The epsilon can be set based on the interquantile distance values obtained from the data, or expert knowledge may provide this value.

## Fuzzy LCS

Following the Zadeh's Fuzzy set definition, Fuzzy Sets $C$ (clusters $C$) of sequence $A$ with a length $T$, characterized by membership function $\mu_{c,k}$ where $c=1,...,nc$, denotes Fuzzy Set $C_c$, and $k$ is the sequence (observation) index where $k=1,...,T$. This set $C$ is more often represented by a cluster prototype (center) which can be calculated by means of FCM algorithm. After these prototypes and membership functions are obtained, these sets can be converted into linguistic labels (such as "increasing", "high", etc.) easily.

Crisp LCS algorithm regards the two observations similar if the difference between them is less than an epsilon value, $|a_i - b_j| < \in$. Fuzzy LCS algorithm changes this evaluation to one of the fuzzy calculations, for example, $\max\{\min\{\mu^A_{i,k}, \mu^B_{i,l}\}\} > \alpha - cut$, where $\alpha - cut$ is the threshold value that specifies the similarity of observations, $\mu^A_{i,k}$ is the membership values of $k^{th}$ (k=1,..,number of observation in $A$) observation of sequence $A$ to cluster $i$ (i=1,..,nc) and similarly $\mu^B_{i,l}$ is the membership values of $l^{th}$ (l=1,..., number of observation in $B$) observation of sequence $B$ to $i^{th}$ cluster with $|k - l| \leq \delta$. After these modifications, Fuzzy LCS can be written as:

$$FLCS_{\delta,\in}(A,B) = \begin{cases} 0 & \text{if } A \text{ or } B \text{ is empty} \\ 1 + FLCS_{\delta}(Head(A), Head(B)) & \text{if } \max\{\min\{\mu^A_{i,k}, \mu^B_{i,l}\}\} > \alpha - cut \text{ and } |k-l| \leq \delta \\ \max(FLCS_{\delta,\in}(Head(A),B), FLCS_{\delta,\in}(A,Head(B))) & \text{otherwise} \end{cases} \quad (5)$$

and $Head(A) = (a_1, a_2,..., a_{nd_1-1})$, $Head(B) = (b_1, b_2,..., b_{nd_2-1})$.

The matching based on maximum of the minimum calculation, $\max\{\min\{\mu^A_{i,k}, \mu^B_{i,l}\}\} > \alpha - cut$, can be changed to any fuzzy number matching procedure. In this paper some of the experiments are designed such that the above algorithm is modified as $sum\{\min\{\mu^A_{i,k}, \mu^B_{i,l}\}\} > threshold$.

The proposed algorithm steps can be listed as:

**Step 1:** Initialize, $nc, m, \delta_l, \delta_u$

**Step 2:** $\mu_A \leftarrow FCM(A, nc, m)$, $\mu_B \leftarrow FCM(B, nc, m)$

**Step 3:** do $k \leftarrow k+1$
  $l \leftarrow k - \delta_l$
  do $l \leftarrow l+1$
    Obtain LCS table with fuzzy operators (Equation 5)
  until $l > k + \delta_u$
until $k > \min(i, j)$

**Step 4:** Calculate the distance based on $FLCS_{\delta,\in}(A,B)$

## Experiments

In order to assess the performance of the Fuzzy LCS, we present several examples including the real world time series data. These experiments are designed in such a way that they include random sequences, deterministic sequences and real world data.

### Random Walk

An important example of random sequences is known as random walk. It is given as:

$x_t = x_{t-1} + \varepsilon_t$ where $\varepsilon_t$ i.i.d. with $N(0, \sigma^2)$

It is known that two random walks may contain stochastic trends that may be correlated even though error terms are uncorrelated. In order to find out how Fuzzy LCS algorithm performs with random walks, a simulation experiment is designed. For this purpose, two random walks are obtained with the following parameters:

$x_n = x_0 + \sum_{i=1}^{n} \varepsilon_i$ where $\varepsilon_i$ i.i.d. with $N(0,1)$

$y_n = y_0 + \sum_{j=1}^{n} \varepsilon_j^*$ where $\varepsilon_j^*$ i.i.d. with $N(0,1)$ and $Cov(\varepsilon_i^*, \varepsilon_i) = 0$ for $i = 1,...,n$

and the parameters of Fuzzy LCS are set as:

$\delta_u = \delta_l = 12$, $\alpha = 0.8$ and number of clusters, $nc = 2,...,9$ for $n \in \{50, 100, 500, 1000, 2000\}$

The Fuzzy LCS is calculated for the differences of each random walk (by matching $(x_k - x_{k-1})$ and $(y_{k'} - y_{k'-1})$ where $\delta_l \leq k - k' \leq \delta_u$ (since the difference of the series used, the clusters obtained may be classified as, *increasing*, *flat*, *decreasing*, etc.). Figure 1 shows mean Fuzzy LCS calculated with 5000 random walk pairs for each case. Following Ozkan and Turksen (2007), three levels of fuzziness, 1.4, 2 and 2.6, are used and each box represents these values respectively. Sequence lengths (n) are 50, 100, 500, 1000 and 2000 as shown with different point shapes given at the right of the Figure 1. It can be seen from the Figure 1 that as the both number of clusters and the levels of fuzziness increase, Fuzzy LCS algorithm produces smaller values. Moreover, the change in Fuzzy LCS values with respect to the change in number of clusters gets smaller for the higher values of number of clusters. Furthermore, Fuzzy LCS algorithm produces quite similar values for every sequence lengths. These values become a bit visible for the cases calculated with the upper bound value of the level of fuzziness. Figure 2 shows the ranges of mean values together with the confidence intervals defined as ±2 standard deviations. These intervals decrease with increasing in both the sequence length and the level of fuzziness. Some selected quantiles of Fuzzy LCS values are given in Table 2. in the Appendix 1.

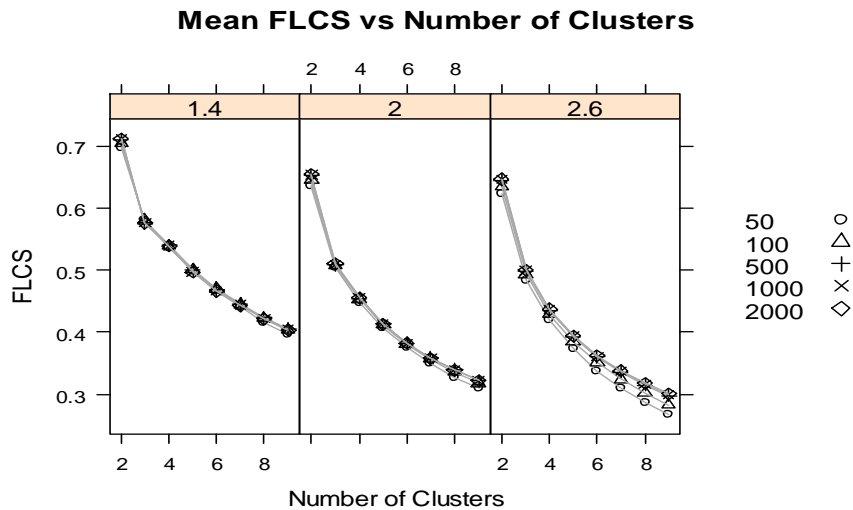

*Figure 1. Mean Fuzzy LCS estimations.*

As a summary, Fuzzy LCS values for random walk pairs are: i) decrease with increase in both the number of clusters and the level of fuzziness, ii) do not change significantly with the sequence length and as a final observation, iii) confidence intervals of the estimated means get narrower as both the sequence length and the level of fuzziness increase.

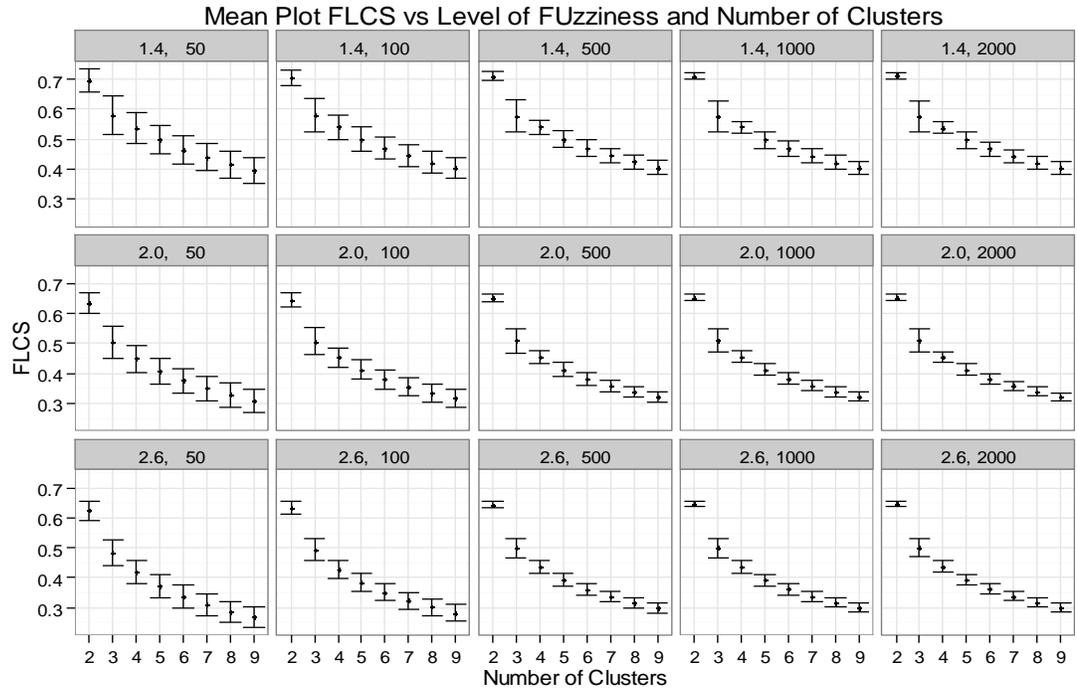

*Figure 2. Mean Fuzzy LCS with confidence intervals.*

*Sine Function*

As a deterministic example we construct a sine and a delayed sine functions as follows;
$x_t = 30 Sin(t)$ and $y_t = x_{t+25}$

To generate these signals one period is divided into 100 equal intervals and totally 125 observation points are used in order to cover one full period. $y_t$ is obtained with $x_t$ by shifting it 25 observations. Function $x_t$ obtained with 30 multiplied by sine function to show the matching delay on the same graph. Figure 3 shows Fuzzy LCS matching between these two sequences. '+' shows the number of lags where values are matched. The time lag between matched points are exactly 25 time steps as expected. In this example, alpha, level of fuzziness, number of clusters are shown at the bottom of the Figure 3. Fuzzy LCS matching is performed on differenced series with the value of the level of fuzziness is two and the number of clusters is 5. Algorithm successfully captures optimal matching points. The similarity calculated by Fuzzy LCS algorithm is approximately 0.798 for this example. There are 125 observations where the first 99 observations of one series are perfectly matched with 26[th] to 125[th] observation of the other series (99/125 is then calculated as Fuzzy LCS similarity).

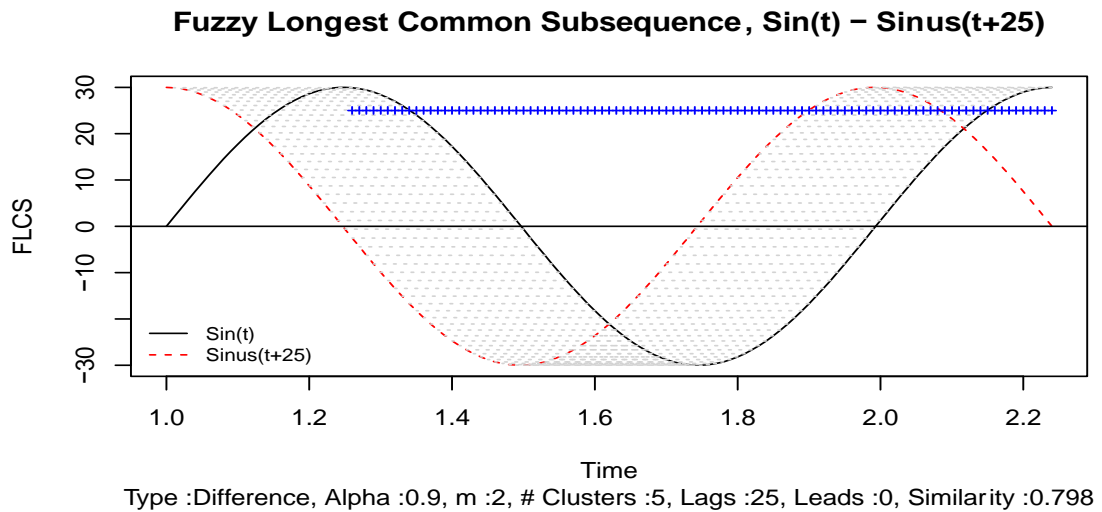

Type :Difference, Alpha :0.9, m :2, # Clusters :5, Lags :25, Leads :0, Similarity :0.798

*Figure 3. Fuzzy LCS for sine and delayed sine functions pair.*

## Currency Examples

Another set of examples are prepared using with real data. Foreign exchange series are among the most researched series in finance. These rates are determined on the foreign exchange markets. Often economists are interested in measuring co-movement between currencies. It is also important to analyze the dynamics of co-movement in time. For example, co-movement during crises, crashes are particularly important for both practitioner and academicians. Figure 4 shows the monthly Australian and New Zealand dollars and their matching points. This example is chosen since these two currencies do move together in general. The period for the analysis is chosen such that it covers the global crisis period (the peak between 2008 and 2010). Maximum leads and lags are set to 12 months. During almost whole period New Zealand dollar leads Australian dollar almost up to eight months. Starting early 2004 to until late 2005, during 2007 and after 2010 their movements are matched within two months. During global crisis, both Australian and New Zealand dollars co-move for six months. During this crisis period of the lead-lag structure of matching disappears. Then around 2010 they started to move together at the same time. The similarity is calculated to be 0.633 in value. The chance of having this value for random walks is negligible as it can be checked from the Table 2. in Appendix 1.

Another currency pair example is given in Figure 5 for Canadian dollar (CAD) and Euro. Their behaviors seem to be similar except some time periods where one of them makes a peak (or bottoms out) before the other one. There are six to ten months of delay between their movements during the peak formation between 2001 and 2002. Euro started to peak its formation before CAD.

One bottom formation appears before global crisis where CAD started to peak before Euro. In this example, the number of clusters is set to 6, which can be seen a bit higher.

The level of fuzziness is set to 1.4 which results in a relatively similar clustering scheme with crisp clustering. Based on the Figure 5, one might want to play with different lead-lag values and other parameters such as alpha-cut, number of clusters and the level of fuzziness. The correlation of differenced currency series is calculated as 0.447. Their Fuzzy LCS similarity is calculated as 0.527. The parameters are set to different values to give another example with different values. According to the Table 2. in the Appendix 1, the chance of having this value with the given parameters is less than 2% (Sequence length is 170, number of clusters is 6 and the level of fuzziness is 1.4) for random walks.

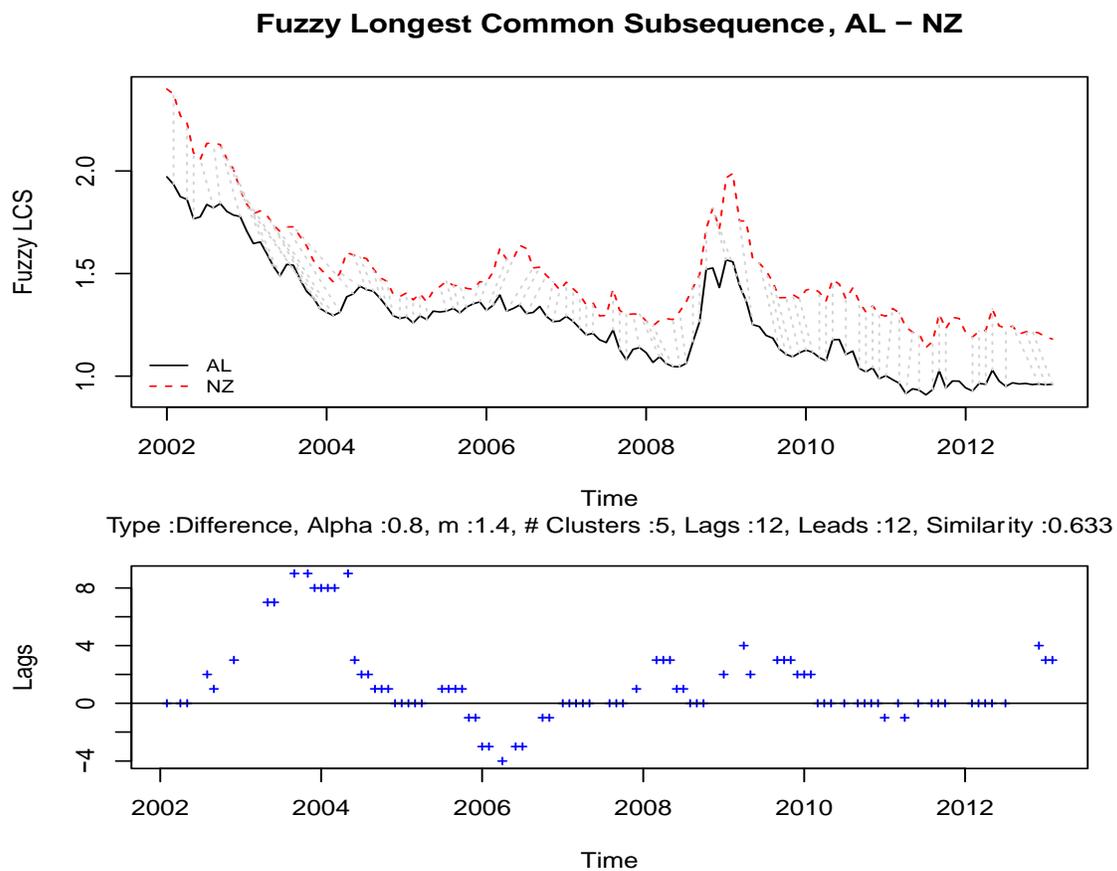

*Figure 4. Fuzzy LCS for Australian and New Zealand dollars.*

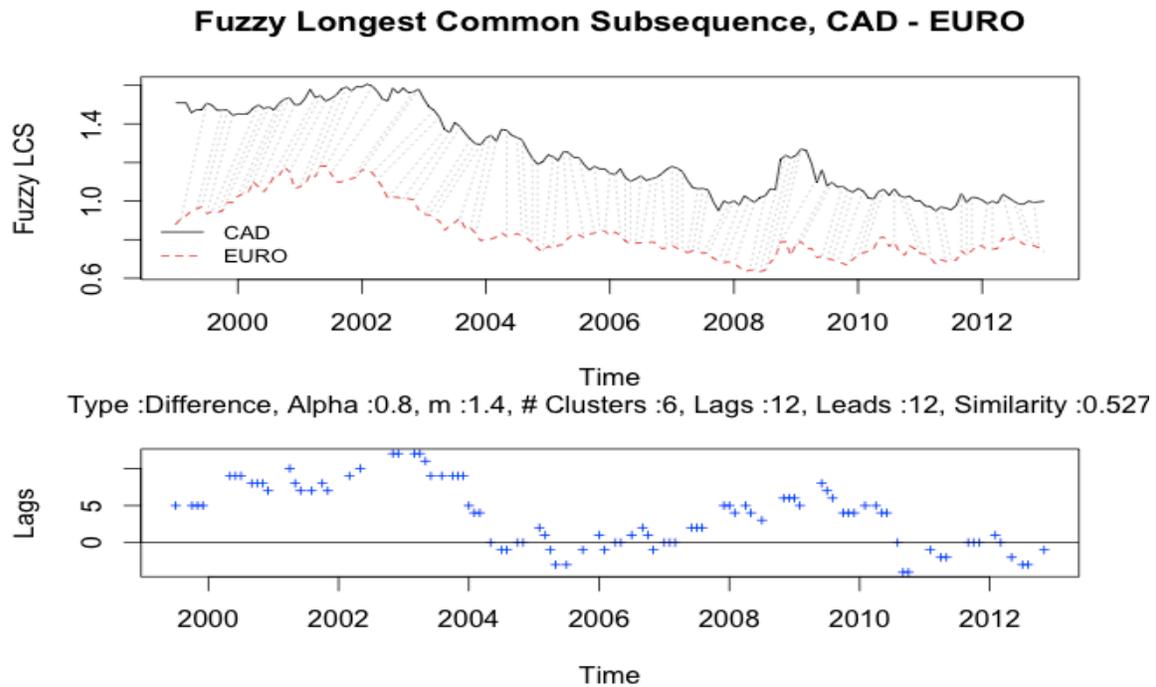

*Figure 5. Fuzzy LCS for Canadian dollar and Euro.*

*Oil Prices and Euro*

We would like to show the application of the Fuzzy LCS to oil price and Euro/$ series as for another example, since the commodity prices are more volatile in general. In addition, the co-movement of exchange rates and oil prices or the dynamics of these series, which are analyzed in the literature since the price of commodity such as oil affect economies (see for example, Sari et al. 2010, Wu et al. 2012). This analysis can be performed by means of Fuzzy LCS to assess the co-movement and/or the inter-relation of both series.

The Fuzzy LCS for weekly Oil prices and €/$ is given in Figure 6. Both these series are scaled to obtain the zero means and unit variances for the sake of presentation and comparability. One can perform the same example with using percentage values as well. In the first step the lead-lag parameters, $\delta_u$ $\delta_l$, are set to 12 to account the effects up to 3 months. The first attempt reveals that the €/$ series lead oil prices. Hence, naturally only lead parameter is set to 12 in the second step. Fuzzy LCS is given in the Figure 6 and Table 1 show the number of weeks between matches together with the number of matches. It seems that approximately 85% and 91.2% of matches already accounted in 6 and 7 weeks delays. The correlation between scaled differenced series are calculated as -0.05438 which is a very small in value. The Fuzzy LCS is calculated as 0.44. Since the number of observations is 710 weeks, number of clusters used in this analysis is 5 and the level of fuzziness is set to 2, Table 2. in Appendix 1 shows that the chance of obtaining Fuzzy LCS as 0.44 is less than 5%. Hence the relation may not be rejected. Another useful information may appear as €/$ series move first then the oil prices follows in many cases.

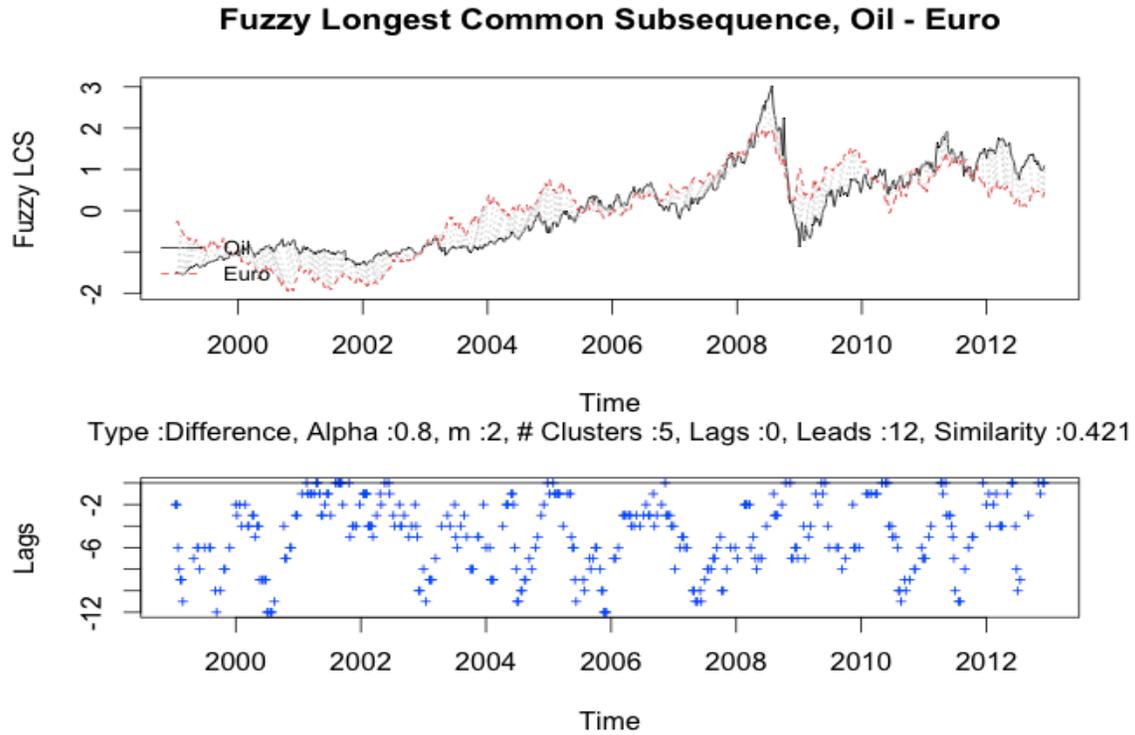

*Figure 6. Fuzzy LCS for Oil and Euro*

Table 1. Oil-Euro Fuzzy LCS matching

| -12 | -11 | -10 | -9 | -8 | -7 | -6 | -5 | -4 | -3 | -2 | -1 | 0 | Number of Weeks Between Matches |
|---|---|---|---|---|---|---|---|---|---|---|---|---|---|
| 8 | 12 | 19 | 20 | 23 | 24 | 37 | 22 | 42 | 34 | 28 | 37 | 30 | Number of Matches |

## CONCLUSION

LCS has been used successfully for different pattern matching problems and similarities between symbolic sequences. Over the last decade, it is observed that there are a use of similarity measures between real valued time series as well. In this paper we propose a novel Fuzzy LCS algorithm with an application of FCM. To our best knowledge, this is the first attempt of constructing Fuzzy LCS with FCM.

In this paper, we provide several examples to show the performance of the Fuzzy LCS. In real world, the observations consist of approximate values. It may be misleading to represent an abstraction of such approximate values based on crisp logic. Therefore, we introduce Fuzzy version of LCS to overcome the chance of obtaining misleading results.

# APPENDIX 1.

Table 2. Fuzzy LCS Simulation Results

Note: 2000 9 2.6 must be read as, sequence length: 2000, Number of Clusters: 9 and Level of Fuzziness: 2.6 Parameters of the simulation is as follows:

Number of Samples for each case: 5000
Sequence Length: 50, 100, 500, 1000, 1500 (5 Cases)
Level of Fuzziness: 1.4, 2, 2.6 (3 Cases)
Number of Clusters: 2, 3, 4, 5, 6, 7, 8, 9 (8 Cases)
Total number of cases: 5*3*8=120

# APPENDIX 2.

```
FLCS <- function (a, mm=2, deltal=12, deltau=12, ty="na",
                            fn=function(a,b) sum(colMins(rbind(a,b))),
                            nc=3, alpha=0.5, ch=1, sc=F, sp=F) {
  myflcs <- list()
  class(myflcs) <- "FLCS"
  a=na.approx(na.trim(zoo(a), side="both"))
  myflcs$m=0
  myflcs$q=a
  myflcs$LCS=0
  myflcs$LCS_opt=c(ch,nc,deltau,deltal,mm,alpha,sc)

names(myflcs$LCS_opt)=c("Opt","NumClust","DeltaLead","DeltaLag","LevelofFuz","alpha","Scale")

  if(sc){
    sa=scale(a)
  } else {
    sa = a
  }

  if (ch > 0){
    da=diff(sa[,1], differences=ch)/ch
    db=diff(sa[,2], differences=ch)/ch
    if (ch ==1 & ty=="per"){
      da=da/sa[1:length(da),1]
      db=db/sa[1:length(da),2]
    }
  } else {
    da=sa[,1]
    db=sa[,2]
  }
  dh=ch

  m = length(da);
  n = length(db);

  if (n<m) {
```

```r
      temp = da
      da = db
      db = temp
      a=a[,c(2,1)]
      temp=deltau
      deltau=deltal
      deltal=temp
      m = length(da)
      n = length(db)
    }

    if (sp){
      centa= min(da)+((1:nc)*((max(da)-min(da))/nc))
      centb= min(db)+((1:nc)*((max(db)-min(db))/nc))

      cl1<-cmeans(da,nc,100,method="cmeans",m=mm)
      s1=sort(cl1$centers,index.return=T)
      cl1$centers=cl1$centers[s1$ix]
      cl1$membership=cl1$membership[,s1$ix]
      myflcs$centersa=as.matrix(cl1$centers)
      U14a=cl1$membership
      myflcs$LCS1_M=U14a

      cl2<-cmeans(db,nc,100,method="cmeans",m=mm)
      s2=sort(cl2$centers,index.return=T)
      cl2$centers=cl2$centers[s2$ix]
      cl2$membership=cl2$membership[,s2$ix]
      myflcs$centersb=as.matrix(cl2$centers)
      U14b=cl2$membership
      myflcs$LCS2_M=U14b
    } else {

      dd=c(as.vector(da),as.vector(db))
      centa= min(dd)+((1:nc)*((max(dd)-min(dd))/nc))
      cl1<-cmeans(dd,nc,100,method="cmeans",m=mm)
      myflcs$centersa=cl1$centers
      myflcs$centersb=cl1$centers
      U14a=cl1$membership[1:m,]
      myflcs$LCS1_M=U14a
      U14b=cl1$membership[(m+1):(m+n),]
      myflcs$LCS2_M=U14b
    }
    lcstable=matrix(0,nrow=(m+1),ncol=(n+1))
    lcstablef=matrix(0,nrow=(m+1),ncol=(n+1))
    prevx=matrix(0,nrow=(m+1),ncol=(n+1))
    prevy=matrix(0,nrow=(m+1),ncol=(n+1))
    mmatch=0

    for (i in 1:m){
      mmatch=0
      for (j in (i-deltal):(i+deltau)) {
        if (j>0 && j<=n){
          kk= fn(U14a[i,],U14b[j,])
          if ( kk > alpha){
            lcstable[i+1,j+1] = lcstable[i,j]+1
            lcstablef[i+1,j+1] = lcstablef[i,j]+kk
            if (kk > mmatch){
```

```
          mmatch= kk
        }
        prevx[i+1,j+1] = i
        prevy[i+1,j+1] = j
      }
      else if (lcstable[i,j+1] > lcstable[i+1,j]) {
        lcstable[i+1,j+1] = lcstable[i,j+1]
        lcstablef[i+1,j+1] = lcstablef[i,j+1]
        prevx[i+1,j+1] = i
        prevy[i+1,j+1] = j+1
      }
      else {
        lcstable[i+1,j+1] = lcstable[i+1,j]
        lcstablef[i+1,j+1] = lcstablef[i+1,j]
        prevx[i+1,j+1] = i+1
        prevy[i+1,j+1] = j
      }
    }
  }
}

lcstable = lcstable[2:(m+1), 2:(n+1)];
myflcs$lcs_table=lcstable

lcstablef = lcstablef[2:(m+1), 2:(n+1)];
myflcs$flcs_table=lcstablef

prevx = prevx[2:(m+1), 2:(n+1)]-1;
prevy = prevy[2:(m+1), 2:(n+1)]-1;

lcs = max(lcstable[m, ]);
lcsf = max(lcstablef[m, ]);
pos= which.max(lcstable[m, ]);
similarity = lcs / (min(m,n));
similarityf = lcsf / (min(m,n));

now = c(m, pos);
prev = c(prevx[now[1], now[2]], prevy[now[1], now[2]]);
lcs_path = now;
while (all(prev>0)){
  now = prev;
  prev = c(prevx[now[1], now[2]], prevy[now[1], now[2]]);
  lcs_path = rbind(lcs_path, now)
}
lcs_path = lcs_path[(dim(lcs_path)[1]):1,]

temp = lcstable[(lcs_path[,2]-1)*m+lcs_path[,1]]; # LCS count along the path
temp = c(0, temp);
index = diff(temp)>0;
match_point = lcs_path[index, ];

if (dim(as.matrix(match_point))[1] > 2){
  myflcs$p_match=match_point
  myflcs$m=1
}
```

```
  myflcs$LCS=c(similarity, similarityf)
  names(myflcs$LCS)=c("LCS","FLCS")
  return(myflcs)
}
```

---

[i] See Ozkan and Turksen 0 for a good survey of Cluster Validity Indices together with their own suggestion